\documentclass[10pt,twocolumn,letterpaper]{article}

\usepackage{cvpr}              

\definecolor{cvprblue}{rgb}{0.21,0.49,0.74}
\usepackage[pagebackref,breaklinks,colorlinks,allcolors=cvprblue]{hyperref}

\def\paperID{*****} 
\def\confName{CVPR}
\def\confYear{2025}

\title{PCIE\_Pose Solution for EgoExo4D Pose and Proficiency Estimation Challenge}

\author{
Feng Chen\\
Lenovo Research\\
{\tt\small chenfeng13@lenovo.com}
\and
Kanokphan Lertniphonphan \\
Lenovo Research\\
{\tt\small klertniphonp@lenovo.com}
\and
Qiancheng Yan\\
University of Chinese Academy of Sciences\\
{\tt\small yanqiancheng21@mails.ucas.ac.cn}
\and
Xiaohui Fan\\
Tsinghua University\\
{\tt\small fxh23@mails.tsinghua.edu.cn}
\and
Jun Xie \\
Lenovo Research\\
{\tt\small xiejun@lenovo.com}
\and
Tao Zhang \\
Tsinghua University\\
{\tt\small taozhang@tsinghua.edu.cn}
\and
Zhepeng Wang \\
Lenovo Research\\
{\tt\small wangzpb@lenovo.com}
}

\begin{document}
\maketitle
\begin{abstract}

This report introduces our team's (PCIE\_EgoPose) solutions for the EgoExo4D Pose and Proficiency Estimation Challenges at CVPR2025. Focused on the intricate task of estimating 21 3D hand joints from RGB egocentric videos, which are complicated by subtle movements and frequent occlusions, we developed the Hand Pose Vision Transformer (HP-ViT+). This architecture synergizes a Vision Transformer and a CNN backbone, using weighted fusion to refine the hand pose predictions. For the EgoExo4D Body Pose Challenge, we adopted a multimodal spatio-temporal feature integration strategy to address the complexities of body pose estimation across dynamic contexts. Our methods achieved remarkable performance: 8.31 PA-MPJPE in the Hand Pose Challenge and 11.25 MPJPE in the Body Pose Challenge, securing championship titles in both competitions. We extended our pose estimation solutions to the Proficiency Estimation task, applying core technologies such as transformer-based architectures. This extension enabled us to achieve a top-1 accuracy of 0.53, a SOTA result, in the Demonstrator Proficiency Estimation competition.

\end{abstract}    
\section{Introduction}
\label{sec:introduction}

Recently, an egocentric video captured using a wearable camera has gained significant importance in the fields of human-computer interaction and robotics. The EgoExo4D Hand Pose challenge, part of the Ego4D Pose benchmark, brings attention to human pose estimation from the egocentric perspective \cite{Grauman2023EgoExo4DUS}. The EgoExo4D Body Pose Challenge aims to accurately estimate body pose using only first-person raw video and/or egocentric camera pose. The input excludes egocentric modalities that would unfairly simplify the task (e.g., audio captured from a wearable camera, eye gaze), as well as exocentric video or any signals that can be extracted from it.

In this work, we address these challenges through the novel frameworks developed for the EgoExo4D Hand Pose and Body Pose Challenges at CVPR2025. :  
\begin{itemize}  
\item \textbf{Hand Pose Estimation:} We introduce the Hand Pose Vision Transformer (HP-ViT+), a hybrid architecture that synergizes the global contextual reasoning of Vision Transformers with the local feature discriminability of CNN backbones. By incorporating weighted fusion mechanisms to refine pose predictions, HP-ViT+ effectively mitigates occlusion issues and captures the intricate kinematic dependencies of 3D hand joints.  
\item \textbf{Body Pose Estimation:} We propose a multimodal spatio-temporal feature integration strategy that fuses head pose sequences, ego-view video features, and depth maps. Leveraging transformer encoders for temporal modeling and cross-modal fusion, this approach capitalizes on complementary cues from multiple modalities to handle the variability of dynamic scenes.  
\item \textbf{Demonstrator Proficiency Estimation:} We further extended our Ego Body Pose framework to the Demonstrator Proficiency Estimation task, which requires classifying the skill level of a demonstrator from synchronized egocentric (wearable camera) and exocentric (third-person) videos of a task performance. Using only egocentric video streams of a demonstrator performing a task, our method achieves the SOTA classification accuracy on the EgoExo4D leaderboard.

\end{itemize}  

\section{Method}
\label{sec:method}

\begin{figure*}[t]
  \centering
   \includegraphics[width=0.9\linewidth]{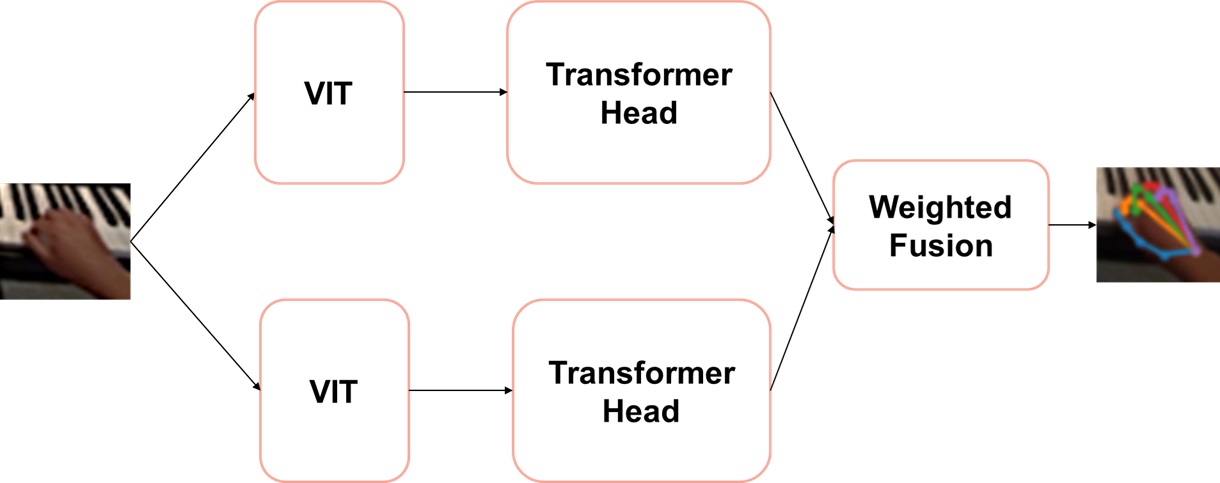}
   \caption{
    The framework of HP-ViT+ for Hand Pose Estimation.
   }
   \label{fig:overview}
\end{figure*}

\subsection{Hand Pose Vision Transformer (HP-ViT+)}

HP-ViT \cite{chen2024pcie_egohandpose} leverages a Visual Transformer (ViT\cite{Dosovitskiy2020Vit}) as the backbone \cite{xu2022vitpose} to extract features from the provided hand image, and a standard transformer decoder \cite{Vaswani2017AttentionIA} locate the hand keypoints. To enhance feature representation capability of HP-ViT, HP-ViT+ additionally integrates ConvNeXt as a complementary CNN backbone. The input hand image $\mathcal{I} \in \mathcal{R}^{H\ast W\ast 3}$ is processed by two parallel pathways, as illustrated by Figure \ref{fig:overview}:

\begin{itemize}  
\item \textbf{ViT Pathway}: the image is divided into 16 X 16 patches $\mathcal{I}_p \in \mathcal{R}^{(H/16)\ast(W/16)\ast3}$, which are fed into a patch embedding layer. The ViT backbone outputs features corresponding to a 1/16 downscaled feature map. 

\item \textbf{ConvNeXt Pathway}: The same input image is processed through the ConvNeXt-Base architecture, which employs hierarchical convolutional layers to extract multi-scale spatial features.
\end{itemize}  

A transformer decoder followed by a MLP head is applied to aggregate the features of human hand pose and predict the 3D keypoint coordinates. The predictions from both ViT and ConvNeXt pathways are then subject to weighted fusion. By leveraging the strengths of both global contextual modeling (ViT) and local spatial detail extraction (ConvNeXt), the fused predictions are computed as a weighted sum based on their respective prediction scores. 

This hybrid architecture (HP-ViT+) significantly improves prediction accuracy by integrating complementary visual cues from both network paradigms.


\begin{figure*}[t]
  \centering
   \includegraphics[width=0.9\linewidth]{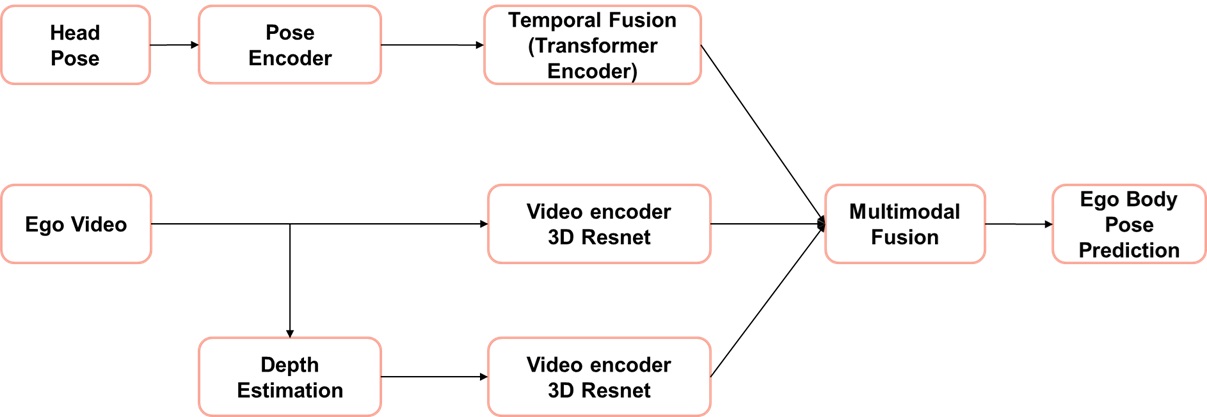}
   \caption{
    Multimodal Spatio-Temporal Fusion Framework for Hand Pose Estimation.
   }
   \label{fig:overview}
\end{figure*}

\subsection{Multimodal Spatio-Temporal Body Pose}

To address the challenges of robust body pose estimation in dynamic ego-centric scenarios, we propose a multimodal network framework that integrates three complementary data streams: \textbf{Head Pose}, \textbf{Ego View Video}, and \textbf{Ego View Depth Estimation}. The architecture is designed to leverage both temporal dynamics and spatial contextual information across modalities, as illustrated in Figure \ref{fig:multimodal_framework}.

\subsubsection{Modal-Specific Encoding}Each input modality is processed by a dedicated encoder to extract discriminative features:

\begin{itemize}  
\item \textbf{Head Pose Modality} Head pose parameters (yaw, pitch, roll) derived from inertial measurement units are encoded using a temporal transformer encoder to capture short-term rotational dynamics. This stream provides prior information about the ego-agent's orientation, crucial for aligning body pose in self-centric coordinates.

\item \textbf{Ego View Video} A sequence of ego-view RGB frames are fed into a 3D CNN encoder (e.g., 3D ResNet \cite{feichtenhofer2019slowfast}) to extract spatio-temporal features $\mathbf{V}_t \in \mathbb{R}^{C_v}$, where $T$ is the temporal window size and $C_v$ is the feature dimension. This captures motion patterns and object interactions in the ego's field of view.

\item \textbf{Ego View Image Depth} Monocular depth estimation from ego-view images (obtained via model DepthAnything2 \cite{yang2024depth}) generates depth-aware embeddings $D \in \mathbb{R}^{T \times C_d}$, which will be encoded by a transformer encoder to produce spatial-temporal depth-aware features $\mathbf{D}_s \in \mathbb{R}^{C_d}$, emphasizing geometric relationships between the body and the environment.
\end{itemize}  

\subsubsection{Spatio-Temporal Fusion}
The encoded features undergo two-stage fusion:

\begin{itemize}  
\item \textbf{Temporal Fusion} For the Head Pose, a transformer encoder is applied to model short-term temporal dependencies. The Head Pose features are first projected to match the other modality feature dimension by frame, after which all frame features are concatenated and fed into the transformer encoder to produce temporally aware embeddings $\mathbf{F}_t \in \mathbb{R}^{C_t}$.

\item \textbf{Spatial Fusion} The fused spatial-temporal representation $\mathbf{F}_{st} \in \mathbb{R}^{C_{st}}$ is then obtained by concatenating the attended head post embeddings, video embeddings and depth embeddings, followed by an MLP encoder layer to model cross-modal interactions.
\end{itemize}  

\subsubsection{Body Pose Prediction}
The unified spatio-temporal features $\mathbf{F}_{st}$ are fed into a shared prediction head consisting of stacked fully connected layers, which regresses 3D body joint coordinates $\hat{\mathbf{Y}} \in \mathbb{R}^{K \times 3}$ (where $K$ is the number of body joints). The loss function employs the Mean Per Joint Position Error (MPJPE) \cite{ionescu2013human3} to measure the Euclidean distance between the predicted and ground truth joints.

This multimodal spatio-temporal framework capitalizes on the complementary strengths of orientation priors (Head Pose), dynamic visual cues (Ego Video), and geometric depth information (Depth Estimation), enabling accurate body pose estimation in challenging ego-centric environments where the ego body is partially or entirely outside the ego view.

\subsection{Demonstrator Proficiency Estimation}

We extend our Ego Body Pose framework to the Demonstrator Proficiency Estimation task, which classifies skill levels from synchronized egocentric (wearable camera) and exocentric (third-person) task videos. To enhance spatio-temporal feature modeling, we integrate a Vision Transformer-based encoder by adopting VideoMAE v2, leveraging its self-supervised pre-training to capture long-range temporal dependencies and fine-grained pose dynamics from first-person viewpoints. We rely solely on egocentric video streams to model complex skill-related spatio-temporal interactions, achieving state-of-the-art (SOTA) classification accuracy on the EgoExo4D benchmark.
\section{Experiment}

\begin{table*}
  \centering
  \begin{tabular*}{\textwidth}{@{\extracolsep{\fill}}
    lcccc} 
    \toprule
     &  \multicolumn{2}{c}{Validation}  &  \multicolumn{2}{c}{Test} \\
    \hline
     \multicolumn{1}{c}{Team} & MPJPE ($\downarrow$) & PA-MPJPE ($\downarrow$) & MPJPE ($\downarrow$) & PA-MPJPE ($\downarrow$)   \\
    \midrule
    HP-ViT+ (Ours)  & \textbf{22.53}  & \textbf{8.54} & \textbf{24.80} & \textbf{8.31} \\
    HP-ViT (Ours)  & 22.67  &  8.59 & 25.51 & 8.49 \\
    Hand3D & - & - & 30.52 &9.30   \\
    Death Knight & - & - & 28.72 & 10.20 \\
    IRMV\_sjtu & - & - & 29.38 & 10.36\\
    Host\_1030\_Team (POTTER manual+auto v2) & - & - & 28.94 & 11.07 \\
    Host\_1030\_Team (POTTER manual v2) & - & - & 30.57 & 11.14 \\
    egoexo4d-hand (POTTER manual+auto v1) & - & - & 29.53 & 11.16 \\
    softbank-meisei (POTTER) & - & - & 62.14 & 19.85 \\
    \bottomrule
  \end{tabular*}
  \caption{EgoExo4D Hand Pose Challenge Leaderboard}
  \label{tab: final}
\end{table*}

\subsection{Dataset and Evaluation metric}

We conducted experiments using the Ego-Exo4D ego pose dataset \cite{Grauman2023EgoExo4DUS}, which contains two separated tasks: hand pose estimation and body pose estimation. 

The \textbf{hand pose estimation} task contains 68K and 340K manual annotations for 2D and 3D, respectively. Additionally, it includes 4.3M and 21M automatic annotations for 2D and 3D. Hand bounding boxes are also provided in the annotation. The Mean Per Joint Position Error (MPJPE) and Procrustes aligned Mean Per Joint Position Error (PA-MPJPE) are used for evaluation. Both metrics are evaluated in the context of the ego-centric coordinate system in millimeters (mm). 

The \textbf{body pose estimation} task is to estimate the 3D human pose sequence using either an egocentric RGB video input, a camera pose sequence, or both. The performance of an algorithm is measured through Mean Per Joint Position Error (MPJPE) in centimeters (cm) and Mean Per Joint Velocity Error (MPJVE) in meters per second (m/s) for the 3D joint positions and velocities, respectively.

The \textbf{Demonstrator Proficiency estimation} task aims to classify a demonstrator's skill proficiency level, using synchronized egocentric and exocentric videos of the demonstrator performing a task. The classification labels include Novice, Early Expert, Intermediate Expert, and Late Expert. Top-1 accuracy is used as the metric. 

\begin{table}
  \centering
  \begin{tabular}{lcc} 
    \toprule
      & MPJPE ($\downarrow$) & PA-MPJPE ($\downarrow$)   \\
    \midrule
    ViT-Base & 24.67 & 9.31\\
    ViT-Large & 23.38 & 9.06 \\
    ViT-Huge & 23.08 & 8.77 \\
    ConvNeXt-V2-Huge & 25.40 & 9.16 \\
    \bottomrule
  \end{tabular}
  \caption{Ablation Study of HP-ViT+ on Hand Pose Estimation }
  \label{tab:ViT size}
\end{table}

\begin{table*}
  \centering
  \begin{tabular*}{\textwidth}{@{\extracolsep{\fill}}
    lcc} 
    \toprule
     & MPJPE ($\downarrow$)  &  MPJVE ($\downarrow$)    \\
    \midrule
    
    Baseline & 18.51  & 0.64  \\
    SJTU-SEIEE & 18.09  & 0.62  \\
    ucb\_ego & 17.19  & 0.57  \\
    zju\_abcd  & 15.32 & 0.55  \\
    EgoCast\_wacv  & 14.36  & 0.59  \\
    head2body  & 12.99  & 0.58  \\
    Multimodal SpatioTemporal Fusion (Ours)  & 11.25  & 0.52  \\
    \bottomrule
  \end{tabular*}
  \caption{EgoExo4D Body Pose Challenge Leaderboard}
  \label{tab:BodyPoseMainResults}
\end{table*}

\subsection{Implementation Detail and Results}

This section describes the experimental setup and results of the proposed hand pose and body pose estimation framework.

\subsubsection{Hand Pose Estimation}

Building on HP-ViT, HP-ViT+ was developed with novel architectural enhancements and ensemble strategies. HP-ViT+ introduced ConvNeXt-V2-Huge \cite{pokhrel2023convnextv2} as a new backbone, reusing HP-ViT’s training configuration: batch size=64, lr=1e-4, MPJPE loss, and 70 epochs of training. The ConvNeXt model achieves 25.40 MPJPE and 9.16 PA-MPJPE in validation set, shown in Table \ref{tab:ViT size}.

During inference, TTA via Horizon Flip was retained. To leverage complementary features, we performed a score-weighted ensemble of ConvNeXt-Huge and Vision Transformer (ViT) outputs, optimized on the validation set. This ensemble strategy reduced MPJPE by 0.71 and PA-MPJPE by 0.18 compared to HP-ViT, achieving 24.8 MPJPE and 8.31 PA-MPJPE, shown in Table \ref{tab: final}, securing the 1st position in the CVPR2025 EgoExo4D Hand Pose challenge. The improvement highlights the synergy between CNN and Transformer architectures in hand pose estimation.


\subsubsection{Body Pose Estimation}

Body pose estimation framework, which employs a multi-modal spatio-temporal fusion architecture. The network takes three input modalities: head pose sequences, ego-view video clips, and ego-view depth maps, and predicts ego body poses through hierarchical spatio-temporal feature fusion.
\paragraph{Model Architecture}
The proposed body pose network processes each modality with dedicated encoders before cross-modal fusion:
\textbf{Head Pose Processing}: A sequence of 16 consecutive camera poses (stride=3) is used as input. These poses are first projected into the video feature space via an MLP projector, followed by a transformer encoder for temporal fusion to capture long-range temporal dependencies in head motion.
\textbf{Ego-View Video Encoding}: A ResNet3D-18 network extracts spatio-temporal features from 4-frame video clips (stride=3), capturing short-range motion cues in the ego-view visual stream.
\textbf{Depth Estimation and Fusion}: Off-the-shelf depth maps are generated using the DepthAnything-v2-Base network for each ego-view frame. The resulting depth sequences undergo temporal fusion via another transformer encoder, modeling depth-related motion patterns.
The features from the three modalities are concatenated and passed through an MLP layer for cross-modal fusion, followed by a regression head to predict the ego body pose.
\paragraph{Training Configuration}
The model is trained with a batch size of 64 using the AdamW optimizer and a cosine annealing learning rate scheduler starting from 1e-4. Training is performed on a single NVIDIA A100 GPU. The loss function is mean per-joint position error (MPJPE) averaged over 17 body joints, following the standard metric in ego-body pose estimation.
\paragraph{Quantitative Results}
Table \ref{tab:BodyPoseMainResults} summarizes the performance on the EgoExo4D body pose dataset. The single-model achieves 11.70 MPJPE (cm) on the validation set and 12.10 MPJPE (cm) on the test set. By training multiple variants with different modal combinations and frame sampling strategies, and ensembling their predictions, the performance is further improved to 10.95 MPJPE (cm) on validation and 11.25 MPJPE (cm) on test, securing the championship in the CVPR2025 EgoExo4D Body Pose Challenge.
\paragraph{Ablation Studies}
Ablation tests (Table \ref{tab:BodyPose_AblationResults}) evaluate the contribution of individual modalities and temporal fusion. Adding any modality (head pose, video, depth) leads to significant performance improvements, confirming the complementarity of multi-modal inputs. Enabling temporal fusion increases MPJPE by 0.4 cm, demonstrating the importance of modeling temporal dynamics. These results validate the design choices of the multi-modal spatio-temporal fusion architecture. 

\subsubsection{Demonstrator Proficiency Estimation}
For the demonstrator proficiency estimation task, we adopt the network framework of Ego Body Pose, using only egocentric video as input. The videoMAE v2 base model serves as the backbone to extract spatio-temporal features from the input videos. A multi-layer perceptron (MLP) head is appended to the backbone to predict the proficiency skill scores.

Following the video configuration of the Body Pose task, we set the stride to 3 and use 8 continuous frames as input for each sample. The training configuration strictly follows that of Ego Body Pose. To enhance the model's performance, we employ an ensemble technique by training multiple models with varied model sizes and frame numbers. 
Our ensemble model achieves a top-1 accuracy of 0.53 on the target dataset, securing the first position on the public leaderboard. The detailed performance comparison with other state-of-the-art methods is presented in Table \ref{tab:Demonstrator_Proficency}.


\begin{table}
  \centering
  \begin{tabular}{lcc} 
    \toprule
     & MPJPE ($\downarrow$)  &  MPJVE ($\downarrow$)    \\
    \midrule
    Camera Pose only & 14.46  & 0.58  \\
    + Ego Camera (1 frame) & 13.04  & 0.67  \\
    + Depth Estimation & 12.42  & 0.60  \\
    + Temporal Fusion  & 12.01 & 0.61  \\
    \bottomrule
  \end{tabular}
  \caption{Ablation Test of Body Pose Estimation}
  \label{tab:BodyPose_AblationResults}
\end{table}

\begin{table}
  \centering
  \begin{tabular}{lc} 
    \toprule
     & Top 1 Accuracy ($\uparrow$)   \\
    \midrule
    Baseline & 14.46  \\
    PCIE Body Pose - vit-b (ours) & 0.51  \\
    PCIE Body Pose - ens (ours) & 0.53 \\

    \bottomrule
  \end{tabular}
  \caption{Demonstrator Proficiency Estimation Leaderboard}
  \label{tab:Demonstrator_Proficency}
\end{table}

    

\section{Conclusion}
\label{sec:conclusion}

This paper presents PCIE\_EgoPose's solutions for CVPR2025 EgoExo4D Pose and Proficiency challenges. For hand pose estimation, the proposed HP-ViT+ integrates Vision Transformer and CNN with weighted fusion, addressing subtle movements and occlusions in egocentric videos to accurately predict 3D hand joints. For body pose estimation, a multimodal spatio-temporal strategy fuses head pose, video, and depth features via transformers, capturing complementary cues for dynamic context modeling. Our methods achieved a championship in both Hand Pose and Body Pose challenges. Additionally, we extended the body pose framework to the Demonstrator Proficiency Estimation task and also won the championship, demonstrating the excellent versatility of our approach. These outcomes validate the effectiveness of transformer-based architectures and cross-modality fusion in overcoming ego-centric perception challenges, highlighting the value of multi-modal integration and temporal modeling for future research in pose estimation.
{
    \small
    \bibliographystyle{ieeenat_fullname}
    \bibliography{main}
}


\end{document}